# Towards Generalizable Trajectory Prediction using Dual-Level Representation Learning and Adaptive Prompting

**Kaouther MESSAOUD**
EPFL
Lausanne, Switzerland
`kaouther.messaoudbenamor@epfl.ch`

**Matthieu CORD**
Valeo.ai
Paris, France
`matthieu.cord@valeo.com`

**Alexandre ALAHI**
EPFL
Lausanne, Switzerland
`alexandre.alahi@epfl.ch`

January 8, 2025

## Abstract

Existing vehicle trajectory prediction models struggle with generalizability, prediction uncertainties, and handling complex interactions. It is often due to limitations like complex architectures customized for a specific dataset and inefficient multimodal handling. We propose **Per**ceiver with **Reg**ister queries (**PerReg+**), a novel trajectory prediction framework that introduces: (1) Dual-Level Representation Learning via Self-Distillation (SD) and Masked Reconstruction (MR), capturing global context and fine-grained details. Additionally, our approach of reconstructing segment-level trajectories and lane segments from masked inputs with query drop, enables effective use of contextual information and improves generalization; (2) Enhanced Multimodality using register-based queries and pretraining, eliminating the need for clustering and suppression; and (3) Adaptive Prompt Tuning during fine-tuning, freezing the main architecture and optimizing a small number of prompts for efficient adaptation. PerReg+ sets a new state-of-the-art performance on nuScenes [1], Argoverse 2 [2], and Waymo Open Motion Dataset (WOMD) [3]. Remarkable, our pretrained model reduces the error by 6.8% on smaller datasets, and multi-dataset training enhances generalization. In cross-domain tests, PerReg+ reduces B-FDE by 11.8% compared to its non-pretrained variant.

## 1 Introduction

Accurate trajectory prediction is essential for intelligent agents, such as autonomous vehicles, to safely and efficiently navigate dynamic, multi-agent environments, directly impacting road safety and traffic flow. However, modeling complex interactions, adapting to diverse scenes, and managing prediction uncertainties remain challenging due to the dynamic nature of real-world driving. Recent studies [4, 5, 6, 7, 8, 9] have introduced self-supervised learning (SSL) in trajectory prediction to improve generalization and performance, using techniques like Contrastive Learning (CL) [4, 10, 8] and Masked Reconstruction (MR) [5, 6, 7] to capture meaningful representations. Additionally, the Perceiver architecture [11, 12], with its efficient processing of multimodal data, has been adapted for trajectory prediction to handle the complexities of real-world inputs.

Despite recent advancements, trajectory prediction methods face several key limitations: *(1) Complex and non-generalizable architectures*—many approaches [13, 5, 14, 15, 16] rely on intricate, highly specialized designs that limit their generalizability across new environments; *(2) Inefficient multimodality handling*—current models [17, 14, 18] often generate numerous trajectory candidates, necessitating computationally intensive clustering, Non-Maximum Suppression (NMS), and ensembling techniques; and *(3) Scalability constraints*—these methods are typically optimized for specific dataset sizes, reducing their effectiveness when scaled to datasets of different sizes. Moreover, the potential of integrating SSL with the Perceiver architecture remains unexplored, potentially limiting the richness of scene representations necessary for robust forecasting across diverse conditions.

To address these limitations and fully harness the potential of both SSL and the Perceiver architecture, we propose a novel approach that integrates SSL techniques—specifically Self-Distillation (SD) [19] and Masked Reconstruction (MR)—within the Perceiver IO architecture for trajectory prediction. These techniques, which are typically applied independently, are combined here to capture both detailed and holistic scene representations crucial for accurate and adaptable forecasting. Through a teacher-student framework, SD aligns *global* representations within the latent space, ensuring consistency even with partial observations, while MR reconstructs masked inputs to facilitate *detailed* learning. To further improve generalization, we leverage the UniTraj framework [20] to pretrain on three diverse trajectory datasets. This multi-dataset pretraining exposes the model to a broad range of driving behaviors and contexts, significantly enhancing its performance and adaptability. Our contributions include:

- **Dual-Level Representation Learning.** We integrate SD and MR to capture both global scene context and fine-grained details, enhancing the model's capability to accurately model complex interactions and scene dynamics.

- **Enhanced Multimodality.** We incorporate register queries [21] alongside mode prediction queries in the decoder, and leverage multimodal prediction during pretraining to maximize the benefits of pretraining tasks like scene reconstruction. By keeping the decoder during finetuning, the model can efficiently manage multimodal predictions without relying on large trajectory pools, clustering, or suppression, resulting in a more effective approach to trajectory forecasting.

- **Adaptive Prompt Tuning.** During fine-tuning, we freeze the main architecture and use a prompt-based strategy with the decoder to optimize a small set of parameters. By utilizing the clustering learned during pre-training, we generate prompts associated with specific clusters, enabling efficient adaptation to new scenarios or datasets without retraining the entire model.

- **Improved Generalization.** Evaluated on nuScenes [1], Argoverse 2 [2], and the Waymo Open Motion Dataset [3], PerReg+ achieves state-of-the-art performance on the UniTraj benchmark. Key results include 6.8% reduction in B-FDE on smaller datasets due to pretraining, improved generalization through multi-dataset training, and an 11.8% reduction in B-FDE in cross-domain tests compared to the non-pretrained variant.

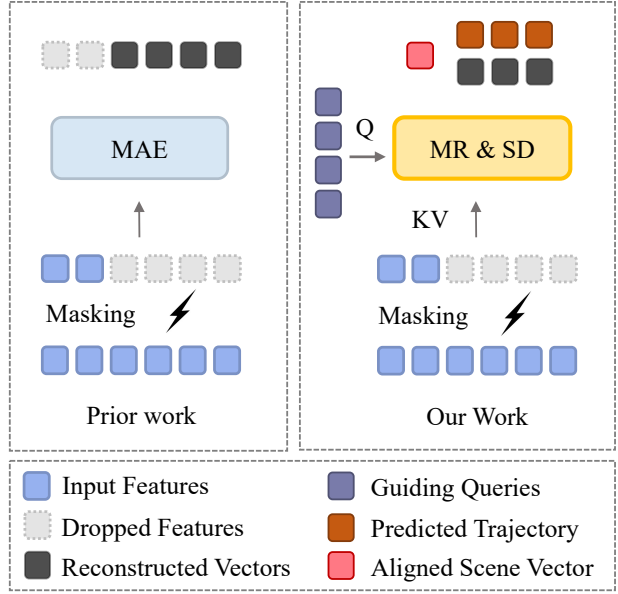

Figure 1: **Comparison between Masked Autoencoder (MAE) and our pretraining strategy with PerReg+.** Our approach incorporates dual-level representation learning through Self-Distillation (SD) and Masked Reconstruction (MR). Unlike MAE, which reconstructs each masked point independently, our method uses segment-level queries to reconstruct entire trajectories (past or future) and lane segments, enabling the decoder to infer complete paths from fine-grained masked inputs. Additionally, we enable multimodal prediction during pretraining, allowing the model to leverage additional insights from the reconstruction process.

Our model demonstrates that integrating a transformer-based architecture with advanced representation learning techniques and large-scale data pretraining leads to richer representations and significant improvements in trajectory prediction accuracy and generalizability.

## 2 Related Work

### 2.1 Trajectory Prediction

**Trajectory Prediction Architectures.** Trajectory prediction in autonomous driving has significantly evolved, utilizing a range of architectures to improve map encoding and interaction modeling [22, 23, 17, 14, 24, 15, 25]. While earlier methods often relied on convolutional neural networks (CNNs) with rasterized imagery [23, 26, 27], newer approaches have shifted toward vectorized data representations using models like transformers [14, 24, 17, 6, 28, 25] and graph neural networks (GNNs) [29, 13, 30, 31]. This progression includes the development of various transformer architectures—such as standard, hierarchical, and spatiotemporal models [32, 13, 25, 33, 34]—and diverse GNN variants [22, 35, 36], enhancing the ability to predict and analyze traffic agent behaviors. However, these sophisticated models, with their increased complexity and large number of parameters, can heighten the risk of overfitting and may limit generalizability across different driving scenarios.

**Multi-modal Trajectory Generation.** In autonomous driving, multi-modal trajectory prediction tackles the uncertainties of dynamic driving conditions by generating multiple potential future paths [22, 23, 17]. Some methods [14, 17] output a fixed set of trajectory proposals and employ a winner-takes-all loss that backpropagates only for the closest match to the observed path. Others expand on this by incorporating predefined trajectories or anchor points [26, 37] and often use a two-stage strategy: first identifying feasible goals within drivable areas, then generating plausible trajectories toward these goals [38, 39]. Alternatively, some methods generate large trajectory pools [14, 17], applying Non-Maximum Suppression (NMS) and clustering to refine predictions. Building on these foundations, we highlight the benefits of using register queries alongside mode prediction queries within the decoder.

### 2.2 Self-supervised Trajectory Prediction

Self-supervised learning (SSL) has shown success in trajectory prediction [4, 5, 6, 7, 8], using pretext tasks to improve model generalizability. We categorize SSL approaches by their masking strategy, representation learning approach, architecture retention, and fine-tuning strategy.

**Masking Strategy.** Masking strategies differ by granularity: fine-grained masking [40, 5, 7] targets individual trajectory and lane points or waypoints, while coarse masking [40, 6] covers entire segments. Fine-grained masking often pairs with fine-grained reconstruction [40, 5, 7], masking each point or segment as a discrete token. Coarse masking typically pairs with coarse reconstruction, learning broader segments (*e.g.,* past or future trajectories [6]). We combine fine-grained masking with coarse reconstruction to capture spatial detail while supporting broader context.

**Holistic vs. Detailed Representation Learning.** In trajectory prediction, holistic learning, primarily using contrastive methods [10, 4, 8], captures high-level semantic relationships by distinguishing similar and dissimilar embeddings. Generative methods [6, 5, 7] reconstruct masked inputs to capture finer spatial and temporal details. Our approach combines both, leveraging generalizable, high-level features while preserving spatial relationships crucial for precise predictions.

**Finetuning the Decoder.** After pretraining, most methods, such as [6, 7, 5, 40], retain only the encoder, discarding the decoder, though adding a randomly initialized decoder during fine-tuning risks making the encoder forgets the pretraining knowledge [41]. To the best of our knowledge only Forecast-PEFT [42] keeps the pretrained decoder during finetuning to preserve learned representations, highlighting the need for fine-tuning strategies that adapt the decoder to multi-modal motion prediction while safeguarding pretraining knowledge.

**Fine-Tuning Strategy.** Most trajectory prediction methods [6, 7, 8] fine-tune the entire network for high accuracy, which is computationally intensive. Parameter-Efficient Fine-Tuning (PEFT) [43, 44], by contrast, updates only a subset of parameters to reduce costs. Recent work [42] uses PEFT to target specific components, essential for resource-limited scenarios. We adopt Prompt Tuning (PT) [43] for faster adaptation while retaining pretraining benefits.

## 3 Methodology

### 3.1 Preliminary

**Problem Formulation.** Predicting the possible future trajectories of an agent in dynamic environments requires multi-modal inputs, specifically its surrounding agents' historical data and road graph information. We define the agents' history tensor $\mathbf{H} \in \mathbb{R}^{A \times T \times D_h}$, where $A$ represents the number of agents, $T$ the number of past time steps, and $D_h$ the dimensionality of state features, such as position $(x, y)$, velocity, and acceleration. The road graph tensor $\mathbf{R} \in \mathbb{R}^{S_r \times P_r \times D_r}$ captures road features represented as polylines, where $S_r$ is the number of road segments, $P_r$ the number of points per segment, and $D_r$ the dimensionality of road features.

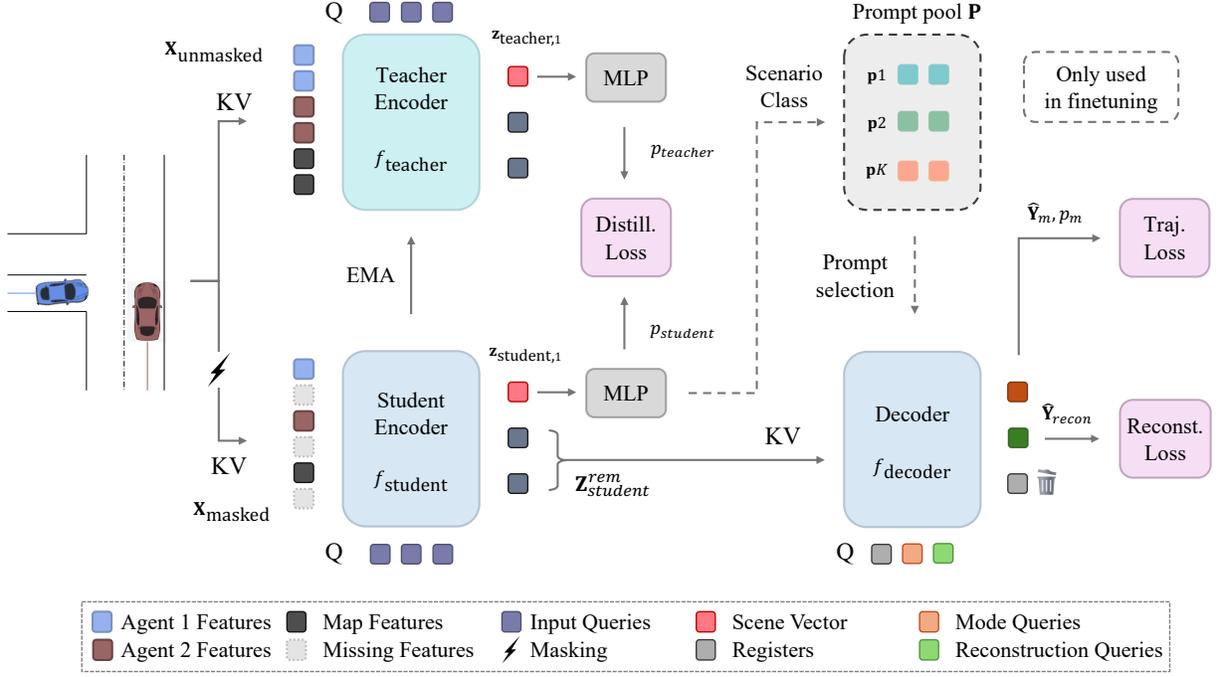

Figure 2: **Overview of the Proposed Trajectory Prediction Model.** The model combines agent history and road information into a unified scene representation, processed by the Perceiver IO architecture to predict future trajectories. In the SD process, teacher and student Perceiver encoders operate in coordination: the teacher encoder (updated via Exponential Moving Average) processes unmasked data with future trajectories, generating context-rich representations to guide the student encoder, which operates on masked inputs. These representations are aligned through a Cross Entropy Loss after the scene clustering head (MLP). The decoder then utilizes the remaining latent representations for trajectory prediction and segment-level scene reconstruction. During fine-tuning, a prompt pool is introduced, with prompts selected dynamically based on scene clustering, allowing efficient scenario-specific adaptation while preserving the pretrained model's representations.

Given inputs $\mathbf{H}$ and $\mathbf{R}$, the task is to learn a mapping function $f$ that predicts the target agent's multimodal future trajectories $\hat{\mathbf{Y}} \in \mathbb{R}^{M \times T' \times D_y}$, where $M$ represents the number of prediction modes, $T'$ the number of future time steps, and $D_y$ the dimensionality of output states (e.g., predicted positions $(x, y)$). This prediction task is formalized as:

$$\hat{\mathbf{Y}} = f(\mathbf{H}, \mathbf{R}). \tag{1}$$

**Scene Representation.** We first encode $\mathbf{H}$ and $\mathbf{R}$ into a unified feature space using separate linear layers to obtain the agent features $\mathbf{A}_{\text{emb}} \in \mathbb{R}^{A \times T \times D}$ and road features $\mathbf{R}_{\text{emb}} \in \mathbb{R}^{S_r \times P_r \times D}$, where $D$ is the embedding dimension. Temporal positional encodings are added to $\mathbf{A}_{\text{emb}}$ to capture sequential dependencies over time, while spatial positional encodings are applied to both $\mathbf{A}_{\text{emb}}$ and $\mathbf{R}_{\text{emb}}$ to preserve spatial structure information. The embedded agent and road graph features are then concatenated along the token dimension, forming a mixed input tensor $\mathbf{X}_{\text{mixed}} \in \mathbb{R}^{N \times D}$, where $N$ is the total number of tokens after concatenation.

**Perceiver IO.** The unified scene representation $\mathbf{X}_{\text{mixed}}$ is processed by the Perceiver IO model, which uses cross-attention and self-attention mechanisms to capture complex spatiotemporal dependencies in high-dimensional data. In the first stage, cross-attention maps the mixed input features $\mathbf{X}_{\text{mixed}} \in \mathbb{R}^{N \times D}$ to a fixed-size latent space $\mathbf{Z}_{\text{latent}} \in \mathbb{R}^{L \times D}$, enabling the model to focus on salient information across modalities:

$$\mathbf{Z}_{\text{latent}} = \texttt{CrossAttention}(\mathbf{Q}_{\text{in}}, \mathbf{X}_{\text{mixed}}), \tag{2}$$

where $\mathbf{Q}_{\text{in}} \in \mathbb{R}^{L \times D}$ denotes learnable latent queries. In the second stage, multiple self-attention layers allow interactions among latent representations $\mathbf{Z}_{\text{latent}}$, enhancing information sharing across modalities and temporal dimensions. Finally, an output querying mechanism extracts scenario-specific information from the latent representations generated by the encoder.

## 3.2 Masked Self-Distillation

To improve robustness in scene encoding and enhance feature learning, we adopt a teacher-student architecture with two Perceiver encoders. The teacher processes unmasked inputs with future trajectories, while the student uses masked inputs. This setup encourages the student model to learn a generalized and context-aware representation from incomplete information.

**Fine-Grained Input Masking.** To promote a comprehensive understanding of the scene, we apply fine-grained masking to input tensors, covering trajectory points and lane polyline points. Unlike previous work that uses coarse masking (where large contiguous regions are masked), fine-grained masking introduces randomness at a finer scale. This approach better mirrors real-world uncertainties, forcing the model to infer missing details based on subtle contextual cues from the agent history and road features.

**Teacher-Student Encoder Architecture.** Let $\mathbf{X}_{\text{unmasked}}$ and $\mathbf{X}_{\text{masked}}$ denote the unmasked and masked input tensors, respectively. The teacher encoder, denoted $f_{\text{teacher}}$, is updated through an Exponential Moving Average (EMA) of the student encoder $f_{\text{student}}$. The teacher encoder receives $\mathbf{X}_{\text{unmasked}}$ to generate a comprehensive scene representation, while the student encoder processes $\mathbf{X}_{\text{masked}}$. By granting the teacher model access to future trajectory information, it generates informative targets that incorporate knowledge of agents' actual future positions and paths. This enables the teacher to produce more accurate and context-aware representations, which the student model can learn from during training. The latent representations from the teacher and student encoders are:

$$
\begin{aligned}
\mathbf{Z}_{\text{teacher}} &= f_{\text{teacher}}(\mathbf{X}_{\text{unmasked}}), \\
\mathbf{Z}_{\text{student}} &= f_{\text{student}}(\mathbf{X}_{\text{masked}}).
\end{aligned}
\tag{3}
$$

## 3.3 Reconstruction and Trajectory Prediction

By combining segment-level reconstruction with multimodal trajectory prediction, the model learns to generate detailed and contextually rich representations beneficial for multimodal prediction.

**Decoder Query Types** To enable the decoder to perform both trajectory prediction and scene reconstruction, we use different types of learned queries. Each query type guides the decoder towards distinct aspects of the input, allowing it to handle multiple tasks in parallel:

*Segment-Level Reconstruction Queries:* These queries guide the decoder in reconstructing masked points within larger segments, such as agent trajectories (both past and future) and lane polyline points. Specifically, two sets of queries, $\mathbf{Q}_{\text{agents}} \in \mathbb{R}^{2 \times A \times D}$, are used to reconstruct past and future trajectories of each agent, while a set of queries $\mathbf{Q}_{\text{road}} \in \mathbb{R}^{S_r \times D}$ is used for reconstructing each lane segment. Positional encodings condition these queries to focus on specific spatial locations, enabling the decoder to infer missing points within each segment based on contextual information from the surrounding unmasked points. To further enhance model robustness, we apply random query dropping during training. By randomly omitting a subset of the reconstruction queries at each training step, the model learns to handle incomplete information, fostering a holistic scene representation.

*Mode Queries for Trajectory Prediction:* A fixed number of mode queries $\mathbf{Q}_{\text{modes}} \in \mathbb{R}^{M \times D}$ are used to predict the future trajectories of the target agent. Each mode query corresponds to a possible future trajectory, allowing the model to capture the multimodal nature of potential outcomes.

*Register Queries:* These queries, denoted $\mathbf{Q}_{\text{reg}} \in \mathbb{R}^{N_R \times D}$, serve as a structured memory for the decoder, storing intermediate scene representations and agent states. This design enables the model to retain knowledge of learned scene dynamics, which is particularly beneficial during fine-tuning.

**Scene Decoding Output.** The decoder produces two main outputs:

1. *Predicted Future Trajectories and Mode Probabilities:* For each mode query, the decoder outputs a predicted trajectory and its associated probability, modeling the multimodal distribution of possible futures. We use a Gaussian Mixture Model (GMM) to represent the probability distribution across different modes.

2. *Reconstructed Past Trajectories and Lane Features:* The decoder reconstructs the masked past trajectories and lane features, encouraging accurate recovery of the missing fine-grained details.

Formally, the decoder output is defined as:

$$\left\{ (\hat{\mathbf{Y}}_m, \hat{p}_m) \right\}_{m=1}^M, \quad \hat{\mathbf{Y}}_{\text{recon}} = f_{\text{decoder}} \left( \mathbf{Z}_{\text{student}}^{\text{rem}}, \mathbf{Q}_j \right), \quad j \in \{\text{modes}, \text{agents}, \text{road}, \text{reg}\}, \quad (4)$$

where $\hat{\mathbf{Y}}_m$ is the predicted trajectory for mode $m$, $\hat{p}_m$ is the predicted probability of mode $m$, and $\hat{\mathbf{Y}}_{\text{recon}}$ includes the reconstructed past trajectories and lane features. Here, $f_{\text{decoder}}$ denotes the decoder network and $\mathbf{Z}_{\text{student}}^{\text{rem}}$ denotes the remaining latent vectors from the student encoder after extracting the first one for SD.

## 3.4 Loss Functions

To train the model effectively, we define several loss functions for the different tasks.

**Self-Distillation Loss.** The first latent vector from the student encoder, $\mathbf{z}_{\text{student},1}$, is mapped through the scene clustering head (MLP) to produce logits. These are compared to the corresponding teacher output vector using a cross-entropy loss:

$$\begin{aligned} p_{\text{student}} &= \texttt{Softmax} \left( \texttt{MLP} \left( \mathbf{z}_{\text{student},1} \right) \right), \\ p_{\text{teacher}} &= \texttt{Softmax} \left( \texttt{MLP} \left( \mathbf{z}_{\text{teacher},1} \right) \right), \\ \mathcal{L}_{\text{distill}} &= \texttt{CrossEntropy} \left( p_{\text{student}}, p_{\text{teacher}} \right), \end{aligned} \quad (5)$$

where $p_{\text{student}}$ and $p_{\text{teacher}}$ denote the softmax probability distributions of the first vectors from the student and teacher, respectively. This loss encourages the student to align its representations with the teacher's unmasked outputs.

**Multimodal Trajectory Prediction Loss.** We model the distribution of future trajectories using a GMM. For each mode $m$, we predict the Gaussian parameters (mean $\hat{\mu}_m$ and covariance $\hat{\Sigma}_m$) and mode probability $\hat{p}_m$. The GMM loss is defined as:

$$\mathcal{L}_{\text{GMM}} = -\log \left( \sum_{m=1}^M \hat{p}_m \cdot \mathcal{N} \left( \mathbf{Y} \mid \hat{\mu}_m, \hat{\Sigma}_m \right) \right), \quad (6)$$

where $\mathbf{Y}$ is the ground truth future trajectory.

**Reconstruction Loss.** For masked agent trajectories and lane polyline points, we apply an L2 reconstruction loss:

$$\mathcal{L}_{\text{recon}} = \sum \left\| \mathbf{Y}_{\text{GT}} - \hat{\mathbf{Y}}_{\text{recon}} \right\|_2^2, \quad (7)$$

where $\mathbf{Y}_{\text{GT}}$ denotes the ground truth elements (it could be masked or unmasked).

**Total Loss.** The total loss is a weighted sum of the individual losses:

$$\mathcal{L}_{\text{total}} = w_{\text{distill}} \cdot \mathcal{L}_{\text{distill}} + w_{\text{GMM}} \cdot \mathcal{L}_{\text{GMM}} + w_{\text{recon}} \cdot \mathcal{L}_{\text{recon}}. \quad (8)$$

To balance the diverse losses, we employ a Dynamic Weighted Aggregation (DWA) strategy, where the weights $w_i$ for each loss $\mathcal{L}_i$ are dynamically adjusted throughout training (details provided in the supplementary materials).

## 3.5 Fine-Tuning with Prompt-Based Clustering

Unlike previous approaches [6, 7, 40] that discard the decoder after pretraining, we retain the decoder along with the mode and register queries, allowing for efficient adaptation to new scenes while preserving the model's rich scene representation capabilities. During fine-tuning, we freeze the parameters of the main architecture and we introduce a prompt-based fine-tuning strategy to adapt the model to specific scenarios or datasets.

We leverage the scene clustering head, an MLP trained with cross-entropy loss during pretraining (see Equation 5) as a clustering mechanism. This head outputs class logits that implicitly cluster input sequences based on their learned representations, grouping sequences with similar characteristics. Additionally, we create a pool of prompt sequences $\mathbf{P} = [\mathbf{p}_1, \mathbf{p}_2, \dots, \mathbf{p}_K]$, where each prompt sequence $\mathbf{p}_k$ corresponds to a specific cluster $k$, and $K$ is the total number of clusters. Based on the clustering results, These prompts act as cluster-specific embeddings, enabling the model to flexibly adapt to varying scenarios or datasets. During fine-tuning, we optimize only the prompts in the prompt pool and the prediction head, keeping the rest of the architecture frozen. For an input $\mathbf{X}$ with a selected prompt $\mathbf{p}_k$, the model's output is:

$$\hat{\mathbf{Y}} = f_{\text{frozen}}(\mathbf{X}, \mathbf{p}_k), \tag{9}$$

where $\hat{\mathbf{Y}}$ is the predicted future trajectories and $f_{\text{frozen}}$ is the frozen PerReg. By focusing optimization on these prompts, we achieve efficient adaptation without modifying the foundational representations learned during pretraining. (Details provided in the supplementary materials)

Table 1: **Quantitative results.** Performance metrics are presented for *Single-Dataset Training*, where each model is trained and evaluated on the same dataset, and *Multi-Dataset Training*, where models are trained on all datasets together. Lower values indicate better performance across all metrics.

| Dataset | Method | Pre-training | Single-Dataset Training | | | | Multi-Dataset Training | | | |
|---|---|---|---|---|---|---|---|---|---|---|
| | | | B-FDE ↓ | minADE ↓ | minFDE ↓ | MR ↓ | B-FDE ↓ | minADE ↓ | minFDE ↓ | MR ↓ |
| nuScenes | AutoBot [24] | − | 3.36 | 1.21 | 2.62 | 0.40 | 3.07 | 1.12 | 2.24 | 0.36 |
| | MTR [14] | − | 2.86 | 1.06 | 2.33 | 0.40 | **2.27** | 0.85 | 1.81 | 0.32 |
| | Forecast-MAE [6] | − | 2.88 | 1.02 | 2.26 | 0.38 | 2.40 | 0.85 | _1.75_ | _0.26_ |
| | Forecast-MAE [6] | + | _2.81_ | _0.99_ | _2.17_ | _0.36_ | 2.39 | _0.84_ | _1.75_ | _0.26_ |
| | **PerReg (Ours)** | − | 3.06 | 1.04 | 2.50 | 0.42 | 2.32 | _0.84_ | 1.76 | 0.27 |
| | **PerReg (Ours)** | + | **2.62** | **0.93** | **1.97** | **0.32** | _2.28_ | **0.79** | **1.64** | **0.25** |
| AV2 | AutoBot [24] | − | 2.51 | 0.85 | 1.70 | 0.27 | 2.54 | 0.86 | 1.73 | 0.27 |
| | MTR [14] | − | 2.08 | 0.85 | 1.68 | 0.30 | **1.99** | 0.82 | 1.61 | 0.28 |
| | Forecast-MAE [6] | − | _2.07_ | _0.75_ | _1.46_ | **0.20** | 2.04 | **0.74** | _1.44_ | **0.20** |
| | Forecast-MAE [6] | + | **2.05** | **0.74** | **1.43** | **0.19** | 2.04 | **0.74** | _1.44_ | **0.19** |
| | **PerReg (Ours)** | − | 2.38 | 0.85 | 1.75 | 0.28 | 2.12 | _0.76_ | 1.51 | 0.23 |
| | **PerReg (Ours)** | + | _2.07_ | 0.77 | _1.46_ | 0.21 | _2.02_ | **0.74** | **1.41** | **0.19** |
| WOMD | AutoBot [24] | − | 2.47 | _0.73_ | 1.65 | _0.25_ | 2.47 | _0.74_ | 1.66 | 0.25 |
| | MTR [14] | − | 2.13 | 0.78 | 1.78 | _0.22_ | 2.13 | 0.78 | 1.78 | 0.33 |
| | Forecast-MAE [6] | − | 2.36 | 0.76 | 1.75 | 0.28 | 2.31 | _0.74_ | 1.69 | 0.27 |
| | Forecast-MAE [6] | + | 2.30 | 0.75 | 1.75 | 0.28 | 2.29 | _0.74_ | 1.68 | 0.25 |
| | **PerReg (Ours)** | − | _2.10_ | **0.65** | _1.46_ | 0.25 | _2.09_ | **0.65** | _1.45_ | _0.22_ |
| | **PerReg (Ours)** | + | **2.05** | **0.65** | **1.42** | **0.20** | **2.04** | **0.65** | **1.42** | **0.20** |

## 4 Experiments

In this section, we evaluate the performance of our proposed model, PerReg+, across multiple trajectory prediction datasets. We assess the model's effectiveness under both *Single-Dataset Training* and *Multi-Dataset Training* settings to explore its ability to generalize across diverse data distributions. Our experiments include comparisons to recent trajectory prediction models and an in-depth analysis of PerReg+'s out-of-domain generalization capabilities. We also conduct an ablation study to quantify the impact of each key component in the model architecture on prediction accuracy and robustness. More experiments about our model scaling and PT are provided in the supplementary materials.

### 4.1 Experimental Setup

**Datasets.** Throughout the experiments, we are using UniTraj framework [20]. We consider three datasets of varying sizes: nuScenes [1] (32k samples), AV2 [2](180k samples), and WOMD [3](1.8M samples) and we have limited the training and validation samples to vehicle trajectories. The map range extends to a 100m radius. The temporal parameters are set to 2 seconds of historical trajectories and 6 second future trajectories.

**Metrics.** We employ the UniTraj benchmark metrics including brier minimum Final Displacement Error (brier-minFDE) [2], minimum average displacement error (minADE), minimum final displacement error (minFDE), and miss rate (MR), where the number of predicted trajectories in the multimodal setting is 6.

**Implementation Details.** Our model is implemented within the UniTraj framework, using a vanilla Perceiver architecture with the same hyperparameters as the Multi-Axis Wayformer [17] on WOMD, consistently applied across datasets. The model configuration includes a hidden size of 256, intermediate size of 1024, 2 encoder layers, 8 decoder layers, and 192 latent queries, outputting 64 Gaussian Mixture Model (GMM) modes. Training is conducted with the AdamW optimizer, an initial learning rate of 2e-4, and a batch size of 128. Input settings consist of 32 surrounding agents, 256 scene polylines, and 20 points per polyline. We apply masking on 90% of history timesteps, 97% of future timesteps, and 75% of map points. Additionally, we use a query drop ratio of 40%.

Table 2: **Out-of-Domain Generalization.** Evaluation of models trained and fine-tuned on the WOMD dataset and tested on the nuScenes validation data. The results show that PerReg achieves the best out-of-domain generalization, especially when pre-training is applied, demonstrating its ability to generalize to different data distributions.

| Method | Pre-training | Evaluation | | | |
|---|---|---|---|---|---|
| | | B-FDE ↓ | minADE ↓ | minFDE ↓ | MR ↓ |
| AutoBot [24] | – | 3.73 | 1.42 | 2.90 | 0.42 |
| MTR [14] | – | 3.10 | 1.17 | 2.52 | 0.43 |
| Forecast-MAE [6] | – | 3.30 | 1.20 | 2.63 | 0.40 |
| Forecast-MAE [6] | + | 3.28 | 1.19 | 2.60 | 0.39 |
| **PerReg (Ours)** | – | 3.12 | 1.17 | 2.51 | 0.44 |
| **PerReg (Ours)** | + | **2.75** | **1.01** | **2.07** | **0.36** |

## 4.2 Comparison with State-of-the-Art Models

In Table 1, we compare our model, PerReg, and its pre-trained variant (denoted as PerReg+) with recent trajectory prediction models, including AutoBot [24], MTR [14], and Forecast-MAE [6], an SSL-based approach, across the nuScenes, AV2, and WOMD datasets. Results are shown for *Single-Dataset Training*, where models are trained and evaluated on the same dataset, and *Multi-Dataset Training*, where models are trained on all datasets simultaneously. Lower values across all metrics indicate better performance.

In *Single-Dataset Training*, PerReg performs well, particularly with pre-training. For smaller datasets, such as nuScenes and AV2, pre-training improves B-FDE by 11% and 13%, respectively, compared to PerReg, while for the larger WOMD dataset, the improvement is 2.4%. This suggests that pre-training is most beneficial for small datasets, as it enables more robust representation learning.

In *Multi-Dataset Training*, PerReg+ shows substantial improvements across datasets by leveraging the larger data pool for more generalized representations. On WOMD, PerReg+ outperforms models like AutoBot and MTR, and on nuScenes and AV2, it achieves competitive or superior performance, with the best scores on several metrics.

Overall, PerReg+ shows strong adaptability and performance across training settings. Pre-training is crucial for smaller datasets, enhancing representation learning, while Multi-Dataset Training further improves results by using diverse data sources. These findings demonstrate PerReg+'s ability to learn generalized embeddings, achieving competitive results in varied trajectory prediction datasets.

## 4.3 Out-of-Domain Generalization

Table 2 presents the out-of-domain generalization results, with models trained on WOMD and tested on nuScenes. PerReg+ achieves the best performance across all metrics, significantly surpassing Forecast-MAE and other baselines. Unlike Forecast-MAE, which employs full fine-tuning but gains only minor improvements from pre-training (e.g., 0.6% in B-FDE), PerReg+ uses a prompt-based fine-tuning strategy. By freezing the pretrained model and updating only a set of prompts, PerReg+ preserves generalizable features, enabling more effective adaptation to unseen data distributions. This approach leads to an 11.8% reduction in B-FDE compared to its non-pretrained variant, demonstrating PerReg+'s superior cross-domain adaptability for trajectory prediction.

## 4.4 Ablation Study

Table 3 presents an ablation study on PerReg+. Starting from a baseline where all decoder queries are used for prediction and aggregated via NMS, introducing Register Queries (+ Reg) reduces B-FDE by 11% by designating six queries for prediction and the rest as registers. Adding Segment-level Reconstruction (+ SR) and retaining pre-trained decoder during finetuning (+ dec) further enhances scene understanding, lowering B-FDE to 3.02 and 2.97, respectively. Incorporating the multimodal prediction task during pre-training (+ pred) yields a 5.0% improvement, reducing B-FDE to 2.82, while Masked SD (+ MSD) improves learning detailed scene reprentation, lowering B-FDE to 2.76. Prompt Tuning (+ PT) achieves the best performance. Overall, each component contributes incremental improvements, with Register Queries and the multimodal prediction yielding the most significant gains.

**Impact of Reconstruction Query Drop Ratio.** We analyze the effect of the reconstruction query drop ratio on B-FDE for PerReg+, pretrained, fine-tuned, and evaluated on the nuScenes dataset. The query drop ratio specifies the percentage of reconstruction queries randomly omitted during pre-training, encouraging the model to handle incomplete information and reduce overfitting. As shown in Figure 4, increasing the drop ratio up to 40% improves B-FDE, indicating enhanced robustness and generalization. However, ratios above 40% lead to performance declines, with B-FDE notably increasing at 80% due to overly sparse inputs limiting effective reconstruction. At 100% query drop, the model relies solely on MSD and multimodal trajectory prediction, omitting scene reconstruction entirely and resulting in the highest B-FDE. These results suggest that a moderate query drop ratio (around 40%) achieves an optimal balance between robustness and predictive accuracy.

Table 3: **Ablation Study.** Quantitative analysis of the impact of individual components in the PerReg+ model on trajectory prediction performance. The table shows the incremental improvements in the metrics as each module is added: Register-based queries (+Reg), self-distillation (SR), decoder retention (+dec), multimodal prediction (+pred), masked SD (MSD), and prompt tuning (PT). The final PerReg+ model achieves the best results (bold). Grey cells indicate stages without pretraining, and percentages in parentheses represent relative improvements in B-FDE.

| Method | Evaluation | | | |
|---|---|---|---|---|
| | B-FDE ↓ | minADE ↓ | minFDE ↓ | MR ↓ |
| Per | 3.44 | 1.19 | 2.87 | 0.36 |
| + Reg | 3.06 (+11%) | 1.04 | 2.50 | 0.42 |
| + SR | 3.02 (+1.3%) | 1.01 | 2.46 | 0.40 |
| + dec | 2.97 (+1.7%) | 1.00 | 2.42 | 0.38 |
| + pred | 2.82 (+5.0%) | 0.97 | 2.16 | 0.35 |
| + MSD | 2.64 (+6.4%) | 0.95 | 2.01 | 0.35 |
| + PT | **2.62** (+0.8%) | **0.93** | **1.97** | **0.32** |

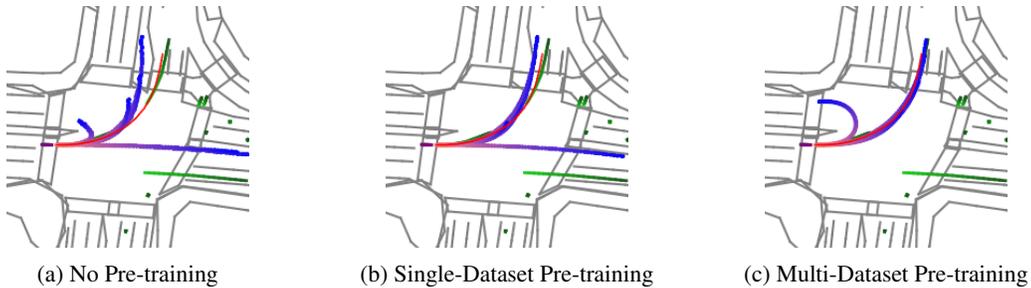

| (a) No Pre-training | (b) Single-Dataset Pre-training | (c) Multi-Dataset Pre-training |

Figure 3: **Comparative Predictions of PerReg in Different Pretraining Settings.** We illustrate three trajectory prediction settings for our model (PerReg) at a complex intersection (nuscenes dataset), showing its performance without pretraining, with single-dataset pretraining, and with multi-dataset pretraining. In each setting, the target vehicle (purple to blue predicted trajectory) is positioned in the leftmost lane with a lead vehicle (green) ahead performing a left turn. Ground truth trajectory is shown in red, with other agents' past trajectories transitioning from light to dark green.

### 4.5 Qualitative Results

We present sample visualizations of our prediction in Figure 3. PerReg predicts possible paths, including a straight path, left turn, and U-turn, but misses following the lead vehicle's left turn. Single-dataset pretraining improves accuracy, showing potential for a left turn but still missing the U-turn. With multi-dataset pretraining, PerReg correctly identifies both the left turn and U-turn possibilities, closely aligning with the lead vehicle's path. Despite the common tendency to continue straight, the model prioritizes turns due to the vehicle's position in the leftmost lane, demonstrating how comprehensive pretraining enhances context-sensitive multimodal predictions.

## 5 Conclusion

We introduced Perceiver with Register queries (PerReg+), a trajectory prediction model addressing key challenges in autonomous navigation. PerReg+ uses Dual-Level Representation Learning (SD and MR) to capture global context and fine-grained details, Segment-Level Reconstruction for trajectories and lane segments to enhance accuracy, and Register-based Multi-Modality to remove clustering requirements for trajectory aggregation. It also features Adaptive Prompt Tuning for efficient fine-tuning. Evaluated on nuScenes, Argoverse 2, and Waymo Open Motion, PerReg+ achieved state-of-the-art results, with pre-training boosting prediction accuracy and cross-domain generalization.

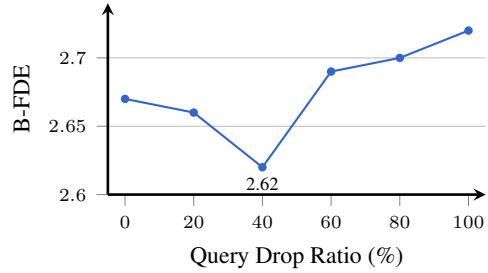

Figure 4: **Impact of Reconstruction Query Drop Ratio.** B-FDE for PerReg+ on the nuScenes dataset across varying reconstruction query drop ratios.

This supplementary document provides additional insights and experiments to complement the main paper. It includes detailed explanations of the Dynamic Weighted Aggregation (DWA) strategy and Prompt-Based Fine-Tuning techniques, along with implementation specifics to ensure consistency in evaluating baselines. Furthermore, we present experiments demonstrating the scalability of our model with increasing data size in multi-dataset training and comparing transfer learning strategies, such as prompt tuning and full fine-tuning, for cross-dataset adaptation.

## 6 Methodology Details

### 6.1 Dynamic Weighted Aggregation (DWA)

To balance the six diverse losses during training, we employ a **Dynamic Weighted Aggregation (DWA)** strategy. Each loss component, denoted as $\mathcal{L}_i$, is assigned a weight $w_i$ that dynamically adjusts based on the relative difficulty of the task at the current stage of training. This approach ensures that harder tasks receive greater emphasis, enabling balanced optimization across all objectives.

The six losses are as follows:

- **Past Reconstruction Loss ($\mathcal{L}_{\text{past}}$):** Reconstruction loss for masked past trajectory points.

- **Future Reconstruction Loss ($\mathcal{L}_{\text{future}}$):** Reconstruction loss for masked future trajectory points.

- **Lane Reconstruction Loss ($\mathcal{L}_{\text{lane}}$):** Reconstruction loss for masked lane polyline points.

- **Prediction Loss ($\mathcal{L}_{\text{pred}}$):** Gaussian Mixture Model (GMM)-based loss to evaluate multimodal trajectory predictions.

- **Cross-Entropy SD Loss ($\mathcal{L}_{\text{cross-entropy}}$):** Aligns the student encoder's latent space with the teacher encoder's outputs.

- **KoLeo Regularization Loss ($\mathcal{L}_{\text{KoLeo}}$) [41]:** This regularizer encourages diversity and uniformity among feature representations within a batch. Given a set of $n$ feature vectors $\{x_1, \ldots, x_n\}$, the regularizer is defined as:

$$\mathcal{L}_{\text{KoLeo}} = -\frac{1}{n}\sum_{i=1}^{n} \log(d_{n,i}), \tag{10}$$

  where $d_{n,i} = \min_{j \neq i} \|x_i - x_j\|$ is the minimum distance between $x_i$ and any other feature in the batch. To ensure consistency and stability, all feature vectors are $\ell_2$-normalized before computing the regularization term.

Each task's loss is dynamically balanced using the DWA. The smoothed loss value $\tilde{L}_i$ for each task $i$ is computed as:

$$\tilde{L}_i = 0.9\, L_i^{(t-1)} + 0.1\, L_i^{(t-2)}, \tag{11}$$

where $L_i^{(t-1)}$ and $L_i^{(t-2)}$ are the loss values from the previous and second-to-last iterations, respectively. The relative importance ratio $r_i$ for task $i$ is then calculated as:

$$r_i = \frac{\tilde{L}_i}{L_i^{(t-2)} + \epsilon}, \tag{12}$$

where $\epsilon$ is a small constant to avoid division by zero. Using these ratios, initial task weights $w_i$ are computed as:

$$w_i = \frac{n\, r_i}{\sum_{j=1}^{n} r_j}, \tag{13}$$

where $n$ is the total number of tasks.

To account for task-specific priorities, biases are applied to the weights, and the biased weights are clipped within predefined bounds $[w_{\min}, w_{\max}]$ to ensure stability. Finally, the weights are normalized again to ensure their sum equals the total number of tasks:

$$w_i = \frac{n\, w_i}{\sum_{j=1}^{n} w_j}. \tag{14}$$

Table 4: **Scalability and Performance Evaluation on Multi-Dataset Training.** PerReg+'s performance is evaluated on combined datasets using data sizes from 20% to 100%, with equal proportions from each, and compared to single-dataset training.

| Data Size | nuscenes | | | | Argoverse 2 | | | | WOMD | | | |
|---|---|---|---|---|---|---|---|---|---|---|---|---|
| | B-FDE ↓ | minADE ↓ | minFDE ↓ | MR ↓ | B-FDE ↓ | minADE ↓ | minFDE ↓ | MR ↓ | B-FDE ↓ | minADE ↓ | minFDE ↓ | MR ↓ |
| 20% | 2.43 | 0.88 | 1.80 | 0.28 | 2.31 | 0.86 | 1.71 | 0.27 | 2.28 | 0.86 | 1.65 | 0.26 |
| 40% | 2.40 | 0.84 | 1.75 | 0.26 | 2.17 | 0.80 | 1.57 | 0.24 | 2.16 | 0.70 | 1.53 | 0.23 |
| 60% | 2.33 | 0.81 | 1.69 | 0.25 | 2.09 | 0.76 | 1.48 | 0.21 | 2.11 | 0.68 | 1.49 | 0.22 |
| 80% | 2.33 | 0.83 | 1.68 | 0.24 | 2.06 | 0.76 | 1.45 | 0.21 | 2.07 | 0.66 | 1.45 | 0.21 |
| 100% | **2.28** | **0.79** | **1.64** | **0.25** | **2.02** | **0.74** | **1.41** | **0.19** | **2.04** | **0.65** | **1.42** | **0.20** |
| Single dataset | 2.62 | 0.93 | 1.97 | 0.32 | 2.07 | 0.77 | 1.46 | 0.21 | 2.05 | **0.65** | **1.42** | **0.20** |

## 6.2 Prompt-Based Fine-Tuning

**Clustering and Prompt Selection**  During fine-tuning, the clustering head, trained during pretraining, assigns each input scene to a specific cluster. The clustering head outputs class logits, and the cluster is identified using the `argmax` operation:

$$\text{cluster\_id} = \arg\max(\text{MLP}(\mathbf{X})), \tag{15}$$

where $\mathbf{X}$ represents the input features, and the cluster with the highest probability is selected. Each cluster corresponds to a unique prompt sequence $\mathbf{p}_k$ from the prompt pool, where $k$ denotes the cluster index.

**Prompt Initialization**  The prompt pool $\mathbf{P} = [\mathbf{p}_1, \mathbf{p}_2, \dots, \mathbf{p}_K]$ consists of $K$ learnable prompt sequences, one for each cluster. Each prompt sequence $\mathbf{p}_k$ is initialized using a uniform distribution:

$$\mathbf{p}_k \sim \mathcal{U}(-1, 1), \quad \forall k \in \{1, 2, \dots, K\}. \tag{16}$$

This initialization ensures diversity across prompt sequences and avoids bias in the adaptation process.

**Integration of Prompts with Queries**  After selecting the prompt sequence $\mathbf{p}_{\text{cluster\_id}}$ based on the assigned cluster, it is concatenated with the mode queries $\mathbf{Q}_{\text{modes}}$ and the register queries $\mathbf{Q}_{\text{reg}}$. The combined query representation $\mathbf{Q}_{\text{combined}}$ is expressed as:

$$\mathbf{Q}_{\text{combined}} = \text{Concat}(\mathbf{Q}_{\text{modes}}, \mathbf{Q}_{\text{reg}}, \mathbf{p}_{\text{cluster\_id}}), \tag{17}$$

where Concat denotes concatenation along the token dimension. This combined query serves as input to the frozen Perceiver decoder.

## 7  Implementation Details

In our implementation, the results for MTR [14] and AutoBot [24] are sourced directly from the UniTraj [20] paper for consistency and comparability. For Forecast-MAE [6], the only SSL-based approach with publicly available code, we adapted the implementation to the UniTraj framework and report the results based on our experiments.

## 8  Additional Experiments

### 8.1  Scalability on Multi-Dataset Training

To evaluate the performance of our model, PerReg+, on multi-dataset pretraining, we combine three datasets of varying sizes and test its performance using progressively larger subsets of the combined data. Specifically, we use 20%, 40%, 60%, 80%, and 100% of the total combined dataset, ensuring that each subset contains equal proportions from all three datasets. This setup allows us to assess how the model scales with increasing data availability while maintaining a balanced representation from each dataset. Performance is evaluated using standard trajectory prediction metrics across these varying dataset sizes.

Table 4 summarizes the results for each dataset and data size. As the size of the combined dataset increases, PerReg+ consistently improves its performance across all metrics, indicating effective utilization of additional data. Notably, the performance at 100% data outperforms the 20% subset by significant margins, particularly in B-FDE and minFDE, reflecting the model's ability to generalize with larger, balanced training data.

When compared to training on single datasets, multi-dataset training with 100% data results in better performance on nuscenes [1] and Argoverse 2 [2] across most metrics, highlighting the benefits of diverse domain exposure. For WOMD [3], results are comparable between the single-dataset and multi-dataset approaches, suggesting that the additional data maintains the model's strong performance. The evaluation demonstrates that multi-dataset training improves the generalization of PerReg+, particularly when leveraging the full combined dataset.

Table 5: **Transfer Pre-training vs. Direct Pre-training.** Evaluation of different pretraining and fine-tuning strategies for the model on the nuscenes dataset. Strategies include Prompt Tuning (PT) on WOMD, nuscenes, and transfer (WOMD → nuscenes), as well as Full Fine-tuning.

| Finetuning Strategy | Evaluation | | | |
|---|---|---|---|---|
| | B-FDE ↓ | minADE ↓ | minFDE ↓ | MR ↓ |
| PT (WOMD) | 2.75 | 1.01 | 2.07 | 0.36 |
| PT (nuscenes) | 2.62 | 0.93 | 1.97 | 0.32 |
| PT (WOMD → nuscenes) | 2.53 | 0.94 | 1.89 | 0.32 |
| Full (WOMD → nuscenes) | **2.27** | **0.79** | **1.64** | **0.27** |

## 8.2 Transfer Pre-training

Table 5 compares various pretraining and fine-tuning strategies on the nuscenes dataset. In all experiments, the model is pretrained on the WOMD dataset and evaluated on nuscenes, except for the PT nuscenes strategy, where the model is both pretrained and fine-tuned exclusively on nuscenes. Prompt tuning on WOMD (PT WOMD) achieves a B-FDE of 2.75, but is outperformed by prompt tuning directly on nuscenes (PT nuscenes). Transfer learning with prompt tuning (PT WOMD → nuscenes) further improves B-FDE, demonstrating the benefits of leveraging large-scale WOMD pretraining. Full fine-tuning after transfer (Full WOMD → nuscenes) achieves the best results across all metrics, highlighting the advantages of fully adapting to the target domain. These results emphasize the importance of combining large-scale pretraining with domain-specific fine-tuning, where prompt tuning offers computational efficiency, and full fine-tuning maximizes performance.